\documentclass{article}
\usepackage{spconf,amsmath,graphicx}
\usepackage{algorithm}
\usepackage{algpseudocode}
\usepackage{csquotes}
\usepackage{multirow}
\usepackage{arydshln}
\usepackage{booktabs}
\usepackage{xcolor}
\usepackage{tikz}

\newcommand{\sta}{\sffamily\footnotesize\textcolor{purple}}
\newcommand{\stb}{\sffamily\footnotesize\textcolor{teal}}
\newcommand{\asr}{\sffamily\footnotesize\textcolor{blue}}
\newcommand{\tsg}{\sffamily\footnotesize\textcolor{darkgray}}

\title{Leveraging Timestamp Information for Serialized Joint Streaming Recognition and Translation}
\name{Sara Papi$^{\ddag}$\sthanks{Work done during an internship at Microsoft.}, Peidong Wang$^\dag$, Junkun Chen$^\dag$, Jian Xue$^\dag$, Naoyuki Kanda$^\dag$, Jinyu Li$^\dag$, Yashesh Gaur$^\dag$}
\address{$^\dag$Microsoft, USA\\
$^\ddag$Fondazione Bruno Kessler and University of Trento, Italy\\
\texttt{\small spapi@fbk.eu,\{peidongwang,junkunchen,jian.xue,nakanda,jinyli,yashesh.gaur\}@microsoft.com}}
\begin{document}
\ninept
\maketitle
\begin{abstract} 
The growing need for instant spoken language transcription and translation is driven by increased global communication and cross-lingual interactions. This has made offering translations in multiple languages essential for user applications.
Traditional approaches to automatic speech recognition (ASR) and speech translation (ST) have often relied on separate systems, leading to inefficiencies in computational resources, and increased synchronization complexity in real time. In this paper, we propose a streaming Transformer-Transducer (T-T) model able to jointly produce many-to-one and one-to-many transcription and translation using a single decoder. We introduce a novel method for joint token-level serialized output training based on timestamp information to effectively produce ASR and ST outputs in the streaming setting. Experiments on \{it,es,de\}$\leftrightarrow$en prove the effectiveness of our approach, enabling the generation of one-to-many joint outputs with a single decoder for the first time.  
\end{abstract}
\section{Introduction}
\label{sec:intro}
With the expansion of global communication and cross-lingual interactions, the demand for real-time spoken language transcription and translation in multiple languages is rapidly increasing \cite{Steigerwald2022OvercomingLB}. 
Conventionally, this task is addressed by separate automatic speech recognition (ASR) and speech translation (ST) models, leading to the necessity of running several models in parallel to obtain the required outputs. This leads to a huge demand for computational resources, in contrast with Green AI \cite{greenai}, and also increases the complexity of coordinating several systems in real-time. 
Moreover, in some applications like news reports, maintaining consistency between on-screen transcriptions and translations
is crucial to deliver to the user similar content \cite{Fuegen2009_1000013594,karakanta-etal-2021-flexibility,xu-etal-2022-joint}. 

Previous works \cite{sperber-etal-2020-consistent,le-etal-2020-dual} have shown improvements in quality and consistency when the model is trained to jointly generate both ASR and ST outputs. This approach was later adapted to the streaming scenario by Weller et al. \cite{weller-etal-2021-streaming} using an attention-based encoder-decoder architecture \cite{10.5555/3295222.3295349} with re-translation \cite{niehues18_interspeech}. More recently, Papi et al. \cite{papi2023token} proposed the adoption of a Transformer-Transducer (T-T) architecture \cite{yeh2019transformer, zhang2020transformer}, which is more suitable for the simultaneous scenario \cite{li2022recent}, with the joint token-level serialized output training (joint t-SOT). 
This method employs an off-the-shelf textual aligner to determine how to effectively produce transcription and translation words in real time. 
However, their study only focused on the many-to-one language setting and can be applied to one translation direction at a time, since finding the alignment between one source language and multiple target languages is a very complex task \cite{pmlr-v89-grave19a,kalinowski2020survey,imani-etal-2022-graph}. 

To overcome this limitation, in this paper, we propose a streaming T-T model that is able to jointly produce both many-to-one and one-to-many outputs using a single decoder. We introduce a novel interleaving method based on timestamp information that enables the model to learn how to produce multiple target languages while maintaining a low latency. Comparative experiments on \{it,es,de\}$\leftrightarrow$en with separate and multilingual state-of-the-art T-T architectures show the effectiveness of our interleaving approach, yielding significant improvements in terms of transcription quality while being competitive when producing multiple translation languages.

\section{Method}
\label{sec:method}

\begin{figure*}
    \centering
    \newcommand{\ImageWidth}{16cm}

\usetikzlibrary{decorations.pathreplacing,positioning, arrows.meta}

\begin{tikzpicture}[x=1cm,scale=1.25]
\draw[thick, -Triangle] (0,0) -- (12.75,0)
node[font=\small,below left=5pt and -8pt]{ms};

\foreach \x/\descr in {0,...,12}
\draw (\x cm,2pt) -- (\x cm,-2pt);

\foreach \x/\descr in {0/0,1/100,2/200,3/300,4/400,5/500,6/600,7/700,8/800,9/900,10/1000,11/1100,12/1200}
\node[font=\small, text height=1.75ex,
text depth=.5ex] at (\x,-.35) {$\descr$};



\node[align=center,above=37.5pt] 
  at (-0.75,0) 
  {\small\sffamily\textbf{ASR:}};
\node[align=center,above=37.5pt] 
  at (2,0) 
  {\asr{I}};
\node[align=center,above=37.5pt,] 
  at (4,0) 
  {\asr{am}};
\node[align=center,above=37.5pt] 
  at (7,0) 
  {\asr{happy.}};

\node[align=center,above=22.5pt] 
  at (-0.75,0) 
  {\small\sffamily\textbf{ST\textsubscript{1}:}};
\node[align=center,above=22.5pt] 
  at (3,0)
  {\sta{Estoy}};
\node[align=center,above=22.5pt] 
  at (9,0) 
  {\sta{feliz.}};

\node[align=center,above=7.5pt] 
  at (-0.75,0) 
  {\small\sffamily\textbf{ST\textsubscript{2}:}};
\node[align=center,above=7.5pt] 
  at (5.2,0) 
  {\stb{Ich}};
\node[align=center,above=7.5pt] 
  at (8,0) 
  {\stb{bin}};
\node[align=center,above=7.5pt] 
  at (11,0) 
  {\stb{froh.}};

\node[align=center,above=52.5pt] 
  at (2.45,0) 
  {\tsg{$t<500$}};
\node[align=center,above=52.5pt] 
  at (7.45,0) 
  {\tsg{$500\leq t<1000$}};
\node[align=center,above=52.5pt] 
  at (11,0) 
  {\tsg{$1000<t\leq1500$}};

\node[align=center,below=30pt] 
  at (5.1,0) 
  {\small\sffamily\textbf{INTER TIME:\qquad\qquad\qquad} \asr{\#ASR\# \textit{I}} \sta{\#ES\# \textit{Estoy}} \asr{\#ASR\# \textit{am}} \stb{\#DE\# \textit{Ich}} \asr{\#ASR\# \textit{happy.}} \stb{\#DE\# \textit{bin}} \sta{\#ES\# \textit{feliz.}} \stb{\#DE\# \textit{froh.}}};
\node[align=center,below=45pt] 
  at (4.5,0) 
  {\small\sffamily\textbf{+ time step grouping:\quad} \asr{\#ASR\# \textit{I am}} \sta{\#ES\# \textit{Estoy}} \stb{\#DE\# \textit{Ich bin}} \asr{\#ASR\# \textit{happy.}} \sta{\#ES\# \textit{feliz.}} \stb{\#DE\# \textit{froh.}}};

\node[rectangle,draw,color=gray,minimum width = 6.1cm, 
    minimum height = 1.6cm,dashed,thick] (r) at (2.45,0.8) {};

\node[rectangle,draw,color=gray,minimum width = 6.1cm, 
    minimum height = 1.6cm,dashed,thick] (r) at (7.45,0.8) {};

\node[rectangle,draw,color=gray,minimum width = 2.5cm, 
    minimum height = 1.6cm,dashed,thick] (r) at (11,0.8) {};


\end{tikzpicture}
    \vspace{-3mm}
    \caption{Timeline visualization of the word sequence for the example in Section \ref{ssec:intertime} with the corresponding {\footnotesize\sffamily \textbf{INTER TIME}} and {\footnotesize\sffamily \textbf{INTER TIME + time step grouping}} joint t-SOT references. English reference ($\mathbf{r_{asr}}$) is {\footnotesize\sffamily \textbf{ASR}}, Spanish reference ($\mathbf{r_{st_1}}$) is  {\footnotesize\sffamily \textbf{ST\textsubscript{1}}}, and German reference ($\mathbf{r_{st_2}}$) is {\footnotesize\sffamily \textbf{ST\textsubscript{2}}}. Time step grouping with time step $T=500ms$ is represented with dashed rectangles.}
    \label{fig:time}
    \vspace{-3mm}
\end{figure*}

\subsection{Joint t-SOT: Review}
\label{ssec:jointtsot}
The joint t-SOT \cite{papi2023token} was proposed to achieve simultaneous ASR and ST by one streaming T-T model \cite{xiechen2021tt}, inspired by t-SOT \cite{kanda22_interspeech} proposed for multi-talker ASR.
Given we have reference transcription 
$\mathbf{r_{asr}}=[r_{asr_1}, ..., r_{asr_M}]$
and 
reference translation
$\mathbf{r_{st}}=[r_{st_1}, ..., r_{st_N}]$ for a training audio,
where $M$ and $N$ are the number of transcription tokens 
and that of translation tokens, respectively.
In the joint t-SOT framework, the T-T model is trained to generate a single sequence of tokens 
including both ASR tokens and ST tokens.
In order to distinguish the tokens for ASR and ST,
two special tokens, $\langle asr \rangle$ and $\langle st \rangle$, 
are inserted in the token sequence like
$[\langle asr \rangle, r_{asr_1},  r_{asr_2}, \langle st \rangle, r_{st_1}, r_{st_2}, \langle asr \rangle, r_{asr_3}, ..., r_{st_N}]$.

Two methods for creating such a serialized token sequence were investigated in \cite{papi2023token}.
The first method was called {\bf INTER $\gamma$} \cite{weller-etal-2021-streaming}, where $\gamma$ can be a real value between 0 to 1.
In the INTER $\gamma$ method, the token sequence is constructed by repeatedly adding either 
transcription tokens or translation tokens starting from $asr_1$ and $st_1$.
In each step, the transcription token is inserted if
$(1 - \gamma) \cdot (1 + asr_i) > \gamma \cdot (1 + st_j)$,
otherwise,
the translation token is inserted.
The special tokens $\langle asr \rangle$ and $\langle st \rangle$ are 
inserted between adjacent transcription tokens and translation tokens.
In case of $\gamma=0.0$ ({\bf INTER 0.0}), 
all transcription tokens will be first generated before the translation tokens.
Contrary, if $\gamma=1.0$ ({\bf INTER 1.0}),
all translation tokens will be first generated before the transcription tokens.
In case of $\gamma=0.5$ ({\bf INTER 0.5}), the transcription and translation tokens are interleaved
in the serialized token sequence.
Among these three variants, only INTER 0.5 is appropriate for simultaneous ASR and ST because 
INTER 0.0 and INTER 1.0 cause a high latency on either ST or ASR.

The second method proposed in \cite{papi2023token} was called
{\bf INTER ALIGN},
where the reference transcription $\mathbf{r_{asr}}$ and
the reference translation $\mathbf{r_{st}}$ are
first aligned by a source-target text alignment tool developed for machine translation.
The alignment forms a bipartite graph between transcription tokens
and translation tokens (see Fig.~2 of \cite{papi2023token}).
We then find disjoint sub-graphs,
each of which represents a pair of transcription tokens and translation tokens.
Finally, the serialized token sequence is created
based on the transcription and translation pairs in the disjoint sub-graphs. 
In \cite{papi2023token}, the INTER ALIGN method achieved 
a better trade-off between accuracy and latency compared to the INTER $\gamma$ method.


\subsection{Timestamp-based joint t-SOT}
\label{ssec:intertime}

While INTER $\gamma$ and INTER ALIGN methods achieved simultaneous ASR and ST, it is not trivial to extend these methods into the multilingual target scenario.
It is especially challenging to extend the INTER ALIGN method for the multilingual scenario because of the difficulty of finding a one-to-many alignment.

To overcome this limitation, in this work, we propose \textbf{INTER TIME}, a novel interleaving method based on word-level timestamps.
The INTER TIME method not only can build more effective joint t-SOT outputs but also enables the one-to-many multilingual scenario by interleaving more than one translation language at a time. 
Note that the usage of the timestamps has been proposed in the t-SOT based multi-talker ASR \cite{kanda22_interspeech} where a phone-based forced aligner \cite{mcauliffe2017montreal} was used to estimate the timestamps. However, it is not possible to follow the same procedure for the ST task where the input audio and output translation are not monotonically aligned.
Therefore, we propose to use model-based emission timestamp. 
Specifically, for each reference word, we compute the corresponding emission timestamp  
by applying the Viterbi algorithm with pretrained streaming models. 
This process is executed for each modality (ASR and ST) and language direction.

In the one-to-many setting, let 
$\mathbf{r_{st_1}}, ..., \mathbf{r_{st_L}}$ be the corresponding translations in $L$ different languages, and
$\langle st_1 \rangle$,..., $\langle st_L \rangle$ be the special tokens indicating the language.\footnote{During training, $\langle asr \rangle$, $\langle st_1 \rangle$,..., $\langle st_L \rangle$ are added to the regular vocabulary and considered in the loss computation as same with all other tokens.}
Each element of $\mathbf{r_{asr}}$, $\mathbf{r_{st_1}}, ..., \mathbf{r_{st_L}}$ is composed of three elements $(time, tag, word)$, where $time$ is the timestamp (integer number, in milliseconds), $tag$ is the corresponding special token (either $\langle asr \rangle$ or $\langle st_l \rangle$), and $word$ is the word that has been emitted with timestamp $time$. For instance, if the word \enquote{\textbf{\textit{I}}} has been uttered at 200ms, the word \enquote{\textbf{\textit{am}}} at 400ms, and the word \enquote{\textbf{\textit{happy.}}} at 700ms (see {\footnotesize\sffamily \textbf{ASR}} in Figure \ref{fig:time}), the corresponding $\mathbf{r_{asr}}$ extracted from the ASR model is:
\begin{gather*}
    \mathbf{r_{asr}} = [(200, \langle asr \rangle, \text{\enquote{\textbf{\textit{I}}}}), (400, \langle asr \rangle, \text{\enquote{\textbf{\textit{am}}}})\\(700, \langle asr \rangle, \text{\enquote{\textbf{\textit{happy.}}}})].
\end{gather*}
If the Spanish translation is \enquote{\textbf{\textit{Estoy feliz.}}} with emission timestamps $[300, 900]$ (see {\footnotesize\sffamily \textbf{ST\textsubscript{1}}} in Figure \ref{fig:time}) and the German translation is \enquote{\textbf{\textit{Ich bin froh.}}} with timestamps $[500, 800, 1100]$ (see {\footnotesize\sffamily \textbf{ST\textsubscript{2}}} in Figure \ref{fig:time}), the corresponding $\mathbf{r_{st_1}}$ and $\mathbf{r_{st_2}}$ extracted from the ST models are:
\begin{gather*}
    \mathbf{r_{st_1}} = [(300, \langle st_1 \rangle, \text{\enquote{\textbf{\textit{Estoy}}}}), (900, \langle st_1 \rangle, \text{\enquote{\textbf{\textit{feliz.}}}})],
\end{gather*}
\begin{gather*}
    \mathbf{r_{st_2}} = [(500, \langle st_2 \rangle, \text{\enquote{\textbf{\textit{Ich}}}}), (800, \langle st_2 \rangle, \text{\enquote{\textbf{\textit{bin}}}})\\(1100, \langle st_2 \rangle, \text{\enquote{\textbf{\textit{froh.}}}})].
\end{gather*}

\begin{algorithm}[t]
\caption{\textsc{INTER TIME}}\label{alg:method}
\begin{algorithmic}
\Require $\mathbf{r_{asr}}$, $\mathbf{r_{st_1}}, ..., \mathbf{r_{st_n}}$ \Comment{ASR and multi ST references}
\State $\mathbf{w} \gets [\quad]$
\For{$r_i$ \textbf{in} $[\mathbf{r_{asr}}$, $\mathbf{r_{st_1}}, ..., \mathbf{r_{st_n}}]$}
\State $\mathbf{w} \gets \mathbf{w} + w_i$ \Comment{Concatenate all the reference words}
\EndFor
\State $\mathbf{w} \gets \text{\textit{sort}}_{time}\big(\mathbf{w}(time, tag, word)\big)$ \Comment{Sort by timestamp}
\State $\mathbf{r_{\text{t-SOT}}} \gets [\quad]$
\State $prev\_tag \gets \text{None}$
\For{$(time, tag, word)$ \textbf{in} $\mathbf{w}$}
\If{$tag \neq prev\_tag$}   \Comment{Language switch}
    \State $\mathbf{r_{\text{t-SOT}}} \gets \mathbf{r_{\text{t-SOT}}} + tag$
    \State $prev\_tag \gets tag$
\EndIf
\State $\mathbf{r_{\text{t-SOT}}} \gets \mathbf{r_{\text{t-SOT}}} + word$
\EndFor
\end{algorithmic}
\end{algorithm}

The INTER TIME output is built by applying Algorithm \ref{alg:method} to $\mathbf{r_{asr}}$, $\mathbf{r_{st_1}}$, $\mathbf{r_{st_2}}$ to obtain the final $\mathbf{r_{\text{t-SOT}}}$. In particular, the reference words for each modality and language are concatenated, sorted by timestamp (increasing order) and then interleaved following the temporal order. 
The special tokens are inserted only if the previous interleaved word was of a different language or domain (ASR or ST).
Following the previous example, the output is:
\begin{align*}
    \mathbf{r_{\text{t-SOT}}} = [&(200, \langle asr \rangle, \text{\enquote{\textbf{\textit{I}}}}), (300, \langle st_1 \rangle, \text{\enquote{\textbf{\textit{Estoy}}}}),\\
    & (400, \langle asr \rangle, \text{\enquote{\textbf{\textit{am}}}}), (500, \langle st_2 \rangle, \text{\enquote{\textbf{\textit{Ich}}}}),\\
    & (700, \langle asr \rangle, \text{\enquote{\textbf{\textit{happy.}}}}), (800, \langle st_2 \rangle, \text{\enquote{\textbf{\textit{bin}}}}),\\
    & (900, \langle st_1 \rangle, \text{\enquote{\textbf{\textit{feliz.}}}}), (1100, \langle st_2 \rangle, \text{\enquote{\textbf{\textit{froh.}}}})],
\end{align*}
with $\langle asr \rangle=\text{\enquote{\#ASR\#}}, \langle st_1\rangle=\text{\enquote{\#ES\#}}, \langle st_2\rangle=\text{\enquote{\#DE\#}}$. The corresponding textual output used during training is shown in Figure \ref{fig:time} ({\footnotesize\sffamily \textbf{INTER TIME}}).
Note that Algorithm \ref{alg:method} can be easily applied to the many-to-one scenario by using different ASR models to obtain multilingual $\mathbf{r_{asr}}$ and a unique $\mathbf{r_{st}}$.

\begin{table*}[!ht]
\setcounter{table}{0}
    \caption{WER$\downarrow$ and BLEU$\uparrow$ on CoVoST 2 for the Many-to-English setting with their ASR latency L$^{\rm ASR}$$\downarrow$ and ST latency L$^{\rm ST}$$\downarrow$. \textbf{Bold} represents overall best result, \underline{underline} represents best result balancing both quality and latency (there can be multiple combinations for each language). }
    \label{tab:multi-source}
    \centering
    \footnotesize
    \setlength{\tabcolsep}{4pt}
    \begin{tabular}{@{}lccccccccccccccc@{}}
    \toprule
       \multirow{2}{*}{Model} &  \multirow{2}{*}{\shortstack[l]{\# inf.\\steps}}  & \multicolumn{4}{c}{it-en} && \multicolumn{4}{c}{es-en} && \multicolumn{4}{c}{de-en} \\
       \cmidrule{3-6} \cmidrule{8-11} \cmidrule{13-16}
       &   & WER & L$^{\rm ASR}$ & BLEU & L$^{\rm ST}$ && WER & L$^{\rm ASR}$ & BLEU & L$^{\rm ST}$ && WER  & L$^{\rm ASR}$ & BLEU & L$^{\rm ST}$ \\
       \midrule
        separate ASR \& ST \cite{papi2023token} & \multirow{2}{*}{6} & 25.83 & 1191 & 16.41 & 1844 && 22.69 & 1149 & 19.24 & 1682 && 23.11 & 1071 & 19.11 & 1613 \\
        multilingual ASR \& ST \cite{papi2023token} &  & 23.48 & 1181 & 21.06 & 1663 && 22.84 & 1147 & 22.76 & 1622 && 21.82 & 1133 & \textbf{21.51} & 1642 \\
        \midrule
        joint t-SOT INTER 0.5 \cite{papi2023token}  & \multirow{2}{*}{3} & 22.35 & 1110 & 20.22 & 1515 && 21.19 & 1126 & 22.25 & 1468 && 21.35 & 1051 & 20.19 & 1547 \\
        joint t-SOT INTER ALIGN \cite{papi2023token} &  & 21.74 & \textbf{1092} & 21.80 & \textbf{1355} && 21.04 & \textbf{1094} & 23.42 & \textbf{1341} && 22.07 & \textbf{1043} & \underline{21.36} & \underline{\textbf{1335}} \\
        \midrule
        joint t-SOT INTER TIME & \multirow{3}{*}{3} & \underline{\textbf{21.11}} & \underline{1141} & 21.70 & 1442 && 19.79 & 1143 & 23.38 & 1452 && \textbf{\underline{21.16}} & \underline{1112} & 19.96 & 1791 \\
        \quad + 500ms step grouping & & 21.22 & 1142 & \underline{\textbf{22.05}} & \underline{1493} && \underline{\textbf{19.74}} & \underline{1139} & \textbf{\underline{24.09}} & \underline{1489} && \underline{21.17} & \underline{1103} & 20.81 & 1664 \\
        \quad + 1000ms step grouping & & 21.64 & 1115 & 21.75 & 1457 && 20.26 & 1052 & 23.75 & 1467 && 21.49 & 1076 & 20.58 & 1651 \\
        \bottomrule
    \end{tabular}
\vspace{-3mm}
\end{table*}

\subsection{Time Step Grouping}
With the aim of limiting the frequency of the switch between languages, we propose the adoption of a grouping mechanism in the data construction process. The grouping mechanism is guided by the size of the time step $T$ (e.g., 500ms, 1000ms, ...). It groups the $(time, tag, word)$ tuple of each reference word $r_i$ of the sorted reference $\text{\textit{sort}}_{time}\big(\mathbf{w}(time, tag, word)\big)$ in Algorithm \ref{alg:method} by looking at the $time$ attribute. Then, the words that belong to the same current time step group $t_s$, i.e. $t_{s}-T \leq time < t_s$, are interleaved together.
The final sequence can be obtained by substituting the $time$ attribute of each word with its corresponding $t_s$ in Algorithm \ref{alg:method}.

For instance, if we look at the example in Section \ref{ssec:intertime} and set the step size $T$ to 500ms, we have three groups $[time<500,500\leq time<1000, 1000\leq time<1500]$, which are visualized as dotted rectangles in Figure \ref{fig:time}. If we substitute $time$ with the corresponding $t_s$ in the sorted $\mathbf{w}$, we obtain:
\begin{align*}
    \mathbf{r_{\text{t-SOT}}} = [&(500, \langle asr \rangle, \text{\enquote{\textbf{\textit{I}}}}), (500, \langle st_1 \rangle, \text{\enquote{\textbf{\textit{Estoy}}}}),\\
    & (500, \langle asr \rangle, \text{\enquote{\textbf{\textit{am}}}}), (1000, \langle st_2 \rangle, \text{\enquote{\textbf{\textit{Ich}}}}),\\
    & (1000, \langle asr \rangle, \text{\enquote{\textbf{\textit{happy.}}}}), (1000, \langle st_2 \rangle, \text{\enquote{\textbf{\textit{bin}}}}),\\
    & (1000, \langle st_1 \rangle, \text{\enquote{\textbf{\textit{feliz.}}}}), (1500, \langle st_2 \rangle, \text{\enquote{\textbf{\textit{froh.}}}})]
\end{align*}
that corresponds to the output {\footnotesize\sffamily \textbf{INTER TIME + time step grouping}} shown in Figure \ref{fig:time}.
The overall language switch reduction ratio depends on the time step $T$. 
In general, the larger $T$ will result in the serialized transcription with less 
special tokens.
On our training data, 
the reduction ratio is estimated as 34\% for 500ms, and 54\% for 1000ms.


\section{Experimental Settings}
\label{sec:exp}
For all our experiments, we use a streaming T-T architecture \cite{xiechen2021tt} with 24 Transformer layers for the encoder with 8 attention heads, 6 LSTM layers for the predictor and 2 feed-forward layers for the joiner. The embedding dimension of the encoder is 512 and the feed-forward units are 4096. We use a chunk size of 1 second with 18 left chunks. The LSTM predictor and feed-forward layers of the joiner have 1024 hidden units. We use 80-dimensional log-mel filterbanks (fbanks) as features, sampled every 10 milliseconds. Before feeding them to the Transformer encoders, we apply 2 layers of CNN with stride 2 and a kernel size of (3, 3), with an overall input compression of 4. The total number of parameters is 188.5M.

Our Many-to-English experiments follow the settings of previous work \cite{papi2023token}: all models are trained for 6.4M steps on 1k hours of proprietary data for each source language (Italian (it), Spanish (es), German (de)) and tested on the CoVoST2 dataset \cite{wang21s_interspeech}. 8k-sized SentencePiece vocabulary \cite{kudo-richardson-2018-sentencepiece} was trained with coverage 1.0 and shared between languages.

For the English-to-Many experiments, we used 1k hours of English audio with the corresponding translation into Italian, Spanish, and German. The models are tested on the FLEURS dataset \cite{fleurs}. 
The multitask multilingual ASR \& ST model is realized by pre-pending the language ID (LID) tag \cite{johnson-etal-2017-googles}, i.e. by replacing the \texttt{<SOS>} with \texttt{<LID>} in the target sequence. Pre-pended LID is used also to train the single-translation version of the joint t-SOT. 
All but separate models are trained for 6.4M steps starting from the multitask multilingual ASR \& ST model weights pretrained for 3.2M steps, including the multitask multilingual model itself. Timestamps for INTER TIME are estimated using monolingual ASR and ST models trained on the same data. Time step grouping is applied at 500ms and 1000ms since preliminary experiments with higher values (e.g., 2000ms) showed quality degradation.

AdamW \cite{loshchilov2018decoupled} is used as optimizer with the RNN-T loss \cite{graves2012sequence}. Checkpoints are saved every 320k steps. The learning rate is set to 3e-4 with Noam scheduler, 800k warm-up steps and linear decay.
We use 16 NVIDIA V100 GPUs with 32GB of RAM for all the training and a batch size of 350k. We select the last checkpoint for inference, which is then converted to open neural network exchange (ONNX) format and compressed. The beam size of the beam search is set to 7. 

We report WER for the ASR quality and BLEU\footnote{sacreBLEU \cite{post-2018-call} version 2.3.1} for the ST quality. Latency is measured in milliseconds~(ms) with the length-adaptive average lagging (LAAL)~\cite{papi-etal-2022-generation}.

\begin{table*}[!t]
\setcounter{table}{1}
    \caption{WER$\downarrow$ and BLEU$\uparrow$ on FLEURS for the English-to-Many setting with their ASR latency L$^{\rm ASR}$$\downarrow$ and ST latency L$^{\rm ST}$$\downarrow$. \textbf{Bold} represents overall best result, \underline{underline} represents best result balancing both quality and latency (there can be multiple combinations for each language). ASR results of joint t-SOT INTER TIME with single translation are averaged among the three languages.}
    \label{tab:multitgt}
    \centering
    \footnotesize
    \setlength{\tabcolsep}{7pt}
    \begin{tabular}{@{}lcccccccccccc@{}}
    \toprule
       \multirow{2}{*}{Model} &  \multirow{2}{*}{\shortstack[l]{\# inf.\\steps}} & \multicolumn{2}{c}{en} && \multicolumn{2}{c}{en-it} && \multicolumn{2}{c}{es-en} && \multicolumn{2}{c}{de-en} \\
       \cmidrule{3-4} \cmidrule{6-7} \cmidrule{9-10} \cmidrule{12-13}
       &  & WER & L$^{\rm ASR}$ && BLEU & L$^{\rm ST}$ &&  BLEU & L$^{\rm ST}$ && BLEU &L$^{\rm ST}$  \\
       \midrule
        separate ASR \& ST & \multirow{3}{*}{4} & 29.02 & 1089 && 8.76 & 1932 && 9.50 & 1853 && 9.87 & 2156 \\
        \quad + multilingual ST &  & 29.02 & 1089 && 11.17 & 1612 && 11.34 & 1618 && \underline{\textbf{13.14}} & \underline{1799} \\
        multitask multilingual ASR \& ST & & 27.53 & 917 && \underline{\textbf{11.56}} & \underline{1607} && 11.38 & 1608 && 13.11 & 1844 \\
        \midrule
        joint t-SOT INTER TIME multi & \multirow{3}{*}{1} & 36.23 & 1544 && 7.52 & 2313 && 8.28 & 2331 && 8.54 & 2497 \\
        \quad + 500ms step grouping & & 31.51 & 1118 && 9.68 & 1668 && 9.86 & 1852 && 11.30 & 1993 \\
        \quad + 1000ms step grouping & & 29.34 & 913 && 10.85 & \textbf{1395} && 10.90 & \textbf{1509} && 12.74 & 1918 \\ \hdashline[1pt/2pt]\hdashline[0pt/1pt] 
        joint t-SOT INTER TIME single & \multirow{3}{*}{3} & \underline{\textbf{26.33}} & \underline{959} && 10.38 & 1564 && 11.24 & 1520 && 12.39 & \textbf{1733} \\
        \quad + 500ms step grouping & & 27.00 & 918 && \underline{11.45} & \underline{1580} && \underline{\textbf{11.89}} & \underline{1610} && 12.79 & 1830 \\
        \quad + 1000ms step grouping & & 26.81 & \textbf{892} && 11.25 & 1776 && 11.52 & 1797 && 12.85 & 1999 \\
    \bottomrule
    \end{tabular}
\vspace{-3mm}
\end{table*}

\section{Results}
\label{sec:results}

\subsection{Many-to-English}
\label{ssec:manytoen}

Table \ref{tab:multi-source} shows the results for the \{it,es,de\}-en language directions. For comparison, we report the results for the joint t-SOT INTER 0.5 and INTER ALIGN approaches. We do not include the results for INTER 0.0 and 1.0 since they are not streaming for one of the two modalities (either ASR or ST).

We first observe that the proposed INTER TIME method without time step grouping achieves higher or similar BLEU scores, except for de-es, and obtains the best WER on all the source languages (it, es, and de). 
The INTER TIME method also achieves comparable ASR and ST latencies with the INTER ALIGN method, which shows
the lowest latency among all methods.
We then observe that the time step grouping improves
overall translation quality while achieving almost the same ASR accuracy and latency scores.
With 500ms time step, the INTER TIME method achieves an average of 0.54 points of BLEU improvement compared to the multilingual ASR \& ST.
It also achieves 0.12 and 0.64 BLEU improvement compared to INTER ALIGN and INTER TIME without time step grouping, respectively. 

All in all, the best quality-latency trade-off is achieved by INTER TIME with time step grouping of 500ms, yielding the best results in most languages and modalities.

\subsection{English-to-Many}

Table \ref{tab:multitgt} shows a comparison
of various ASR \& ST models for English-to-Many setting.
In rows 1-3, we show the results by using 
English ASR model and three ST models (row 1),
one English ASR model and one multi-lingual ASR model (row 2), and one multitask multilingual ASR \& ST model.
Note that, all models in rows 1 to row 3 can output
only one ASR result or ST result at one inference step, 
so we need to execute 4 times of 
inference steps to obtain the listed result.
In this evaluation, we observe that the multitask multilingual ASR \& ST model achieves the best results, with an improved WER and similar or better BLEU scores compared to using separate models for modalities (ASR and ST) and languages.

We then evaluate the proposed joint t-SOT INTER TIME model using multiple translation languages (rows 4-6), or only one translation language (rows 7-9) in the target. The latter model produces the transcription and the corresponding translation in a single language at one inference step, similar to the experiment in Section \ref{ssec:manytoen}.  

For the joint t-SOT INTER TIME model trained on multiple translation languages (rows 4-6), we notice that the time step grouping helps with the performance, both in terms of quality and latency. 
It yields 6.89 points of WER improvement and an average of 3.38 points of BLEU improvement with 983ms latency reduction when the 1000ms of time step grouping is applied. Compared with the strongest system (i.e. multitask multilingual ASR \& ST at row 3), our model shows a marginal degradation of ASR and ST accuracy, with 1.81 point WER degradation and 0.52 point BLEU degradation. 
The latency is slightly improved. 
It is noteworthy that, while marginal degradation of ASR and ST accuracy is observed,
the joint t-SOT INTER TIME model requires only a single inference step.
It is significantly efficient compared to
the multitask multilingual ASR \& ST model that
requires 4 times of inference steps to obtain
all the results. 


If we constrain the joint t-SOT INTER TIME strategy to deal with only one translation language (rows 7-9), we observe significant improvements compared to its multiple translation languages counterpart (rows 4-6), especially with the time step grouping. 
WER improvements range from 2.34 to 3.01 and average BLEU improvements are up to 0.55 compared to the multiple translation model with 1000ms of time step grouping (row 6).  
As was the case with the many-to-English experiment in Section \ref{ssec:manytoen},
the 500ms time step grouping is the best-performing model and, compared with the multitask multilingual ASR \& ST model, it yields 0.53 points of WER improvement while maintaining comparable or slightly better BLEU score and latency.

To conclude, we show the effectiveness of the joint t-SOT INTER TIME, especially when time step grouping is applied. Results on both ASR and ST tasks show that our method achieves the best overall results compared to the strongest multitask multilingual ASR \& ST model. 
When it is extended to deal with multiple translation languages all at once (rows 4-6), our proposed method maintains comparable results while significantly reducing the inference step to only one step.

\section{Conclusions}
In this paper, we proposed a streaming T-T that is able to simultaneously produce many-to-one and one-to-many transcriptions and translations. To effectively train the model to maximize ASR and ST quality while minimizing latency, we proposed INTER TIME, a novel method for the joint t-SOT framework where the tokens are sorted based on the model-based emission timestamp information. We also proposed a variant of this method based on grouping the timestamp according to a fixed time step. 
Comparative studies on \{it,es,de\}-en and en-\{it,es,de\} prove the effectiveness of our approach, especially when the time step grouping is adopted. It achieved the best ASR and ST accuracy in many-to-English ASR and ST
scenario while keeping the ability of low latency inference. 
In English-to-many ASR and ST scenarios, 
the proposed method achieved comparable ASR and ST accuracy
to the baseline model while 
significantly reducing the inference cost.

\bibliographystyle{IEEEbib}
\bibliography{refs}

\end{document}